\documentclass{article}
\PassOptionsToPackage{numbers, sort}{natbib}

\usepackage[eandd, final]{neurips_2026}

\usepackage[utf8]{inputenc} 
\usepackage[T1]{fontenc}    
\usepackage{hyperref}       
\usepackage{url}            
\usepackage{booktabs}       
\usepackage{amsfonts}       
\usepackage{nicefrac}       
\usepackage{microtype}      
\usepackage{xcolor}         
\newcommand{\faded}[1]{\textcolor{black!40}{#1}}

\hypersetup{colorlinks}
\usepackage{booktabs}   
\usepackage{pifont}     
\usepackage[table]{xcolor}
\definecolor{mygray2}{gray}{.6}
\definecolor{mygray3}{gray}{.3}

\definecolor{myblue}{RGB}{0,82,164}
\definecolor{mygreen}{RGB}{0,128,0}

\usepackage{booktabs}
\usepackage{multirow}
\usepackage{multicol}
\usepackage{colortbl}
\usepackage{algorithm}
\usepackage{algpseudocode}
\usepackage{graphicx}
\usepackage{amsmath,amssymb,bm}
\usepackage{subcaption}
\usepackage{enumitem}
\usepackage{wrapfig}

\def\eg{\emph{e.g.}}

\def\ourtask{HieraNav}
\def\ourbenchmark{LangMap}
\def\ourmodel{PlaNaVid}
\def\ourmemory{Bounded Diverse Memory}
\definecolor{lvscene}{RGB}{148,32,147}
\definecolor{lvroom}{RGB}{255,147,0}
\definecolor{lvregion}{RGB}{84,130,53}
\definecolor{lvinst}{RGB}{4,50,255}

\title{LangMap: A Human-Verified Benchmark for Hierarchical Open-Vocabulary Goal Navigation}

\author{
\textbf{Bo Miao}\textsuperscript{\rm 1} \,
\textbf{Weijia Liu}\textsuperscript{\rm 2} \,
\textbf{Jun Luo}\textsuperscript{\rm 1} \,
\textbf{Lachlan Shinnick}\textsuperscript{\rm 1} \,
\textbf{Jian Liu}\textsuperscript{\rm 3} \,
\textbf{Thomas Hamilton-Smith}\textsuperscript{\rm 1} \\
\textbf{Yuhe Yang}\textsuperscript{\rm 4} \,
\textbf{Zijie Wu}\textsuperscript{\rm 4} \,
\textbf{Vanja Videnovic}\textsuperscript{\rm 5} \,
\textbf{Feras Dayoub}\textsuperscript{\rm 1} \,
\textbf{Anton van den Hengel}\textsuperscript{\rm 1,6}
\\[0.1cm]
\textsuperscript{\rm 1} AIML, Adelaide University \;
\textsuperscript{\rm 2} East China Normal University \;
\textsuperscript{\rm 3} SUTD \\
\textsuperscript{\rm 4} The University of Western Australia \;
\textsuperscript{\rm 5} Breaker Industries \;
\textsuperscript{\rm 6} Metacognition AI
\\[0.2cm]
\textbf{Project page:} \href{https://bo-miao.github.io/LangMap/}{bo-miao.github.io/LangMap}
}

\begin{document}

\maketitle

\begin{abstract}
Language-conditioned goal navigation requires embodied agents to locate user-specified targets without step-by-step guidance. However, existing benchmarks largely focus on category-level goals or rely on instance descriptions generated by vision-language models, which often contain ambiguities and semantic errors, limiting systematic and reliable evaluation. We introduce \textbf{\ourtask}, an open-vocabulary navigation task with goals at four hierarchical semantic levels: scene, room, region, and instance. We present Language as a Map (\textbf{\ourbenchmark}), the first goal navigation benchmark to enrich real-world indoor 3D scans with human-verified semantic annotations supporting tasks across all four goal levels. Built on HM3D using a \textit{contrastive annotation protocol} that compares same-scene regions and instances, \ourbenchmark\ provides region labels and discriminative region and instance descriptions covering 414 object categories and contains over 18K tasks. Each target has concise and detailed descriptions, enabling evaluation across instruction styles. Automated and human evaluations validate our annotation quality: our descriptions improve text-to-view matching accuracy over GOAT-Bench's by 23 points on all shared annotated instances, and an independent human audit yields 92.5\% unique-and-correct matches. We also propose \textbf{\ourmodel}, an RGB-only baseline that combines \ourmemory\ with high-level planning to prime a reactive policy for multi-goal navigation, achieving top-tier success rates without depth, 3D scene representations, or object masks. Further analyses reveal that exploration and hierarchical disambiguation failures become more prominent at finer goal levels, while long-tail categories, small objects, distant targets, timely stopping, and multi-goal completion remain challenging.
\end{abstract}

\section{Introduction}

Goal-oriented navigation is fundamental to embodied intelligence, supporting applications such as home-assistant robots. It requires agents to interpret instructions,~\eg, object categories~\cite{ovon,habitatchallenge} or reference images~\cite{krantz2022instance,krantz2023navigating}, and navigate 3D environments to reach targets without step-by-step guidance. We focus on language-conditioned goal navigation (LGN) for intuitive human-robot interaction.

\begin{figure}[t!]
\centering
\begin{minipage}[t]{0.54\linewidth}
    \centering
    \includegraphics[width=\linewidth]{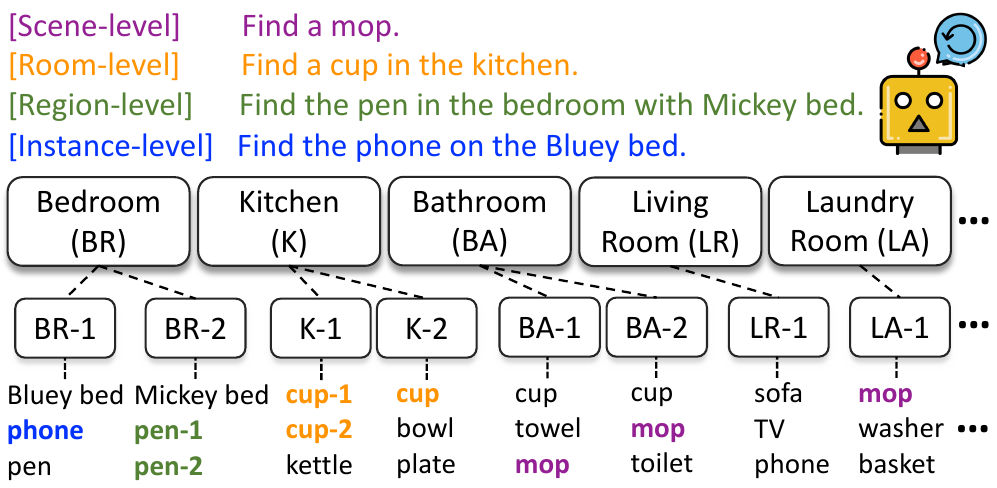}
    \caption{
    \ourtask\ requires agents to navigate to language-specified goals at four semantic levels:
    \textcolor{lvscene}{\textbf{scene}},
    \textcolor{lvroom}{\textbf{room}},
    \textcolor{lvregion}{\textbf{region}}, and
    \textcolor{lvinst}{\textbf{instance}}. Targets are color-coded.
    }
    \label{fig:demo}
\end{minipage}
\hfill
\begin{minipage}[t]{0.43\linewidth}
    \centering
    \includegraphics[width=\linewidth]{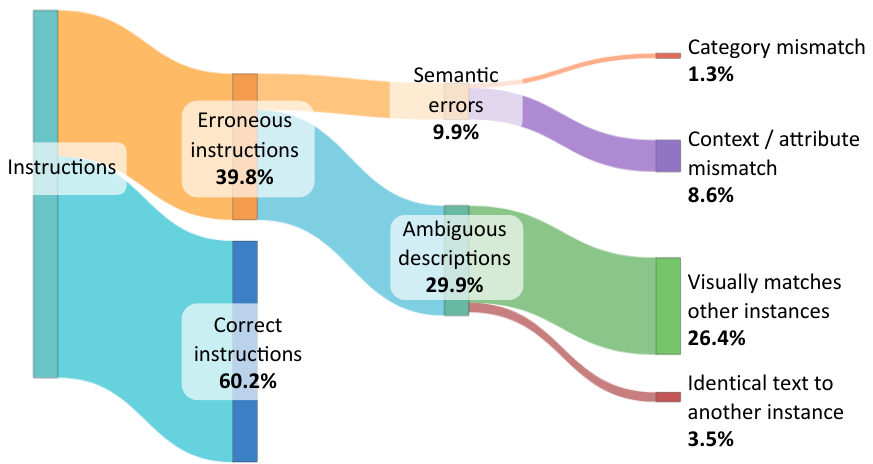}
    \caption{GOAT-Bench annotation quality on ten scenes: 39.8\% of instance-level instructions are erroneous or ambiguous.}\label{fig:goat_error_distribution}
\end{minipage}
\end{figure}

Previous LGN research primarily studies object-goal navigation (ObjectNav), where agents locate \textit{any} instance of a category (\eg, chair). Early benchmarks~\cite{batra2020objectnav,habitatchallenge} adopt a closed-set formulation with 6--21 object categories and evaluate generalization to new environments. Subsequent work extends ObjectNav to open-vocabulary categories~\cite{ovon} and incorporates inferred room types~\cite{lhvln}.
However, category-level navigation prioritizes perception and detection over fine-grained semantic reasoning. 
This is insufficient when users specify context-dependent goals, such as \textit{find the phone on the Bluey bed}, which require both semantic and spatial understanding for disambiguation (see Figure~\ref{fig:demo}).

Recently, GOAT-Bench~\cite{goatbench} and PSL~\cite{PSL} attempt to unify category- and instance-level navigation using vision-language models (VLMs) to generate instructions from HM3D object views~\cite{hm3d,hm3dsem}.
However, VLMs often fail to capture distinctive cues~\cite{li2023evaluating} and spatial relations~\cite{cai2025spatialbot,chen2024spatialvlm}, resulting in descriptions that lack uniqueness and spatial grounding.
Since reliable benchmark evaluation hinges on annotation quality, we manually inspect ten scenes (about 30\% of GOAT-Bench's evaluation set~\cite{goatbench}), examining same-category instance descriptions within each scene for visual grounding and discriminative power.
As shown in Figure~\ref{fig:goat_error_distribution}, 39.8\% of the inspected instance-level descriptions contain \textit{semantic errors} (9.9\%) or \textit{ambiguities} (29.9\%), where nominally ``unique'' descriptions match multiple objects or lack discriminative cues (see Appendix for visualizations).
These findings reveal substantial noise in the VLM-generated annotations and underscore the need for human-verified benchmarks. Beyond these instance-level issues, room- and region-level disambiguation remains largely underexplored in goal-oriented navigation due to the lack of corresponding semantic annotations.

To address these gaps, we argue that a robust LGN benchmark should:
(\textbf{i}) cover diverse goal granularities, from coarse scene-level to fine-grained instance-level;
(\textbf{ii}) provide human-verified, discriminative descriptions that uniquely identify targets within each scene; and
(\textbf{iii}) support open-vocabulary evaluation of both multi-goal episodes spanning mixed semantic levels and single-goal tasks at different levels.
We therefore introduce~\textbf{\ourtask}, a multi-granularity open-vocabulary navigation task that unifies goals across four levels: \textit{scene}, \textit{room}, \textit{region}, and \textit{instance}. 
As illustrated in Figure~\ref{fig:demo}, \ourtask\ captures diverse real-world goal specifications and challenges agents to interpret natural language, perform spatiotemporal reasoning, and navigate to the specified target.

To support rigorous evaluation, we present Language as a Map (\textbf{\ourbenchmark}), a navigation benchmark built on real-world HM3D indoor scans~\cite{hm3d,hm3dsem} and enriched with \textit{comprehensive, human-verified semantic annotations} and \textit{navigation tasks across all four semantic levels}.
Unlike prior navigation datasets, \ourbenchmark\ provides region labels and discriminative region and instance descriptions covering 414 object categories. Of these, 349 categories that pass visibility and viewpoint filtering are used to construct over 18K navigation tasks, providing 2.9$\times$ the category coverage of the VLM-generated annotations in the GOAT-Bench evaluation set~\cite{goatbench}.
Our annotations are produced through a \textit{rigorous contrastive protocol}: annotators compare same-category regions and instances within each scene to write discriminative, natural descriptions, followed by cross-checking for correctness.
Each target is paired with concise descriptions emphasizing salient cues and detailed descriptions providing richer context, enabling evaluation across diverse instruction styles.
Automated and human evaluations validate the quality of~\ourbenchmark: our descriptions improve one-to-many text-to-view matching accuracy over GOAT-Bench's by 23 points on all shared annotated instances, and an independent human audit yields 92.5\% unique-and-correct matches.

We further introduce~\textbf{\ourmodel}, a decoupled RGB-only baseline for multi-goal navigation. It employs~\ourmemory\ (BDM) with a high-level planner to select an initial waypoint and heading for each goal, priming a reactive policy.
\ourmodel\ achieves top-tier success rates on \ourtask\ without depth, 3D scene representations, or object masks.
Our analyses show that exploration and hierarchical disambiguation failures become increasingly prominent at finer goal levels, while long-tail categories, small objects, distant targets, timely stopping, and multi-goal completion remain challenging across methods.

In summary, our main contributions include:

\begin{itemize}
\item
We introduce~\ourtask, an open-vocabulary goal navigation task where agents interpret language instructions to reach target objects specified at four hierarchical semantic levels: scene, room, region, and instance.
\item
We present~\ourbenchmark, the first real-world indoor 3D navigation benchmark with comprehensive human-verified annotations supporting tasks across all four goal levels.
Built with a contrastive annotation protocol, \ourbenchmark\ provides region labels and discriminative region and instance descriptions covering 414 object categories, with 349 categories used to construct over 18K tasks. Each target includes both concise and detailed descriptions.
\item
We propose~\ourmodel, a strong RGB-only baseline that pairs \ourmemory\ with high-level planning to prime reactive navigation, achieving top-tier success rates without depth, 3D scene representations, or object masks.
\item 
We systematically evaluate zero-shot and supervised models on~\ourbenchmark, demonstrating the benefits of memory and richer context and identifying long-tail categories, small objects, distant targets, and multi-goal completion as key challenges. Complementary diagnostics distinguish exploration, hierarchical disambiguation, and termination failures.
\end{itemize}

\section{Related Work}
\label{sec:related_work}

\noindent\textbf{Goal-Oriented Navigation.}
Goal-oriented navigation (GN) enables embodied agents to interpret instructions and navigate 3D environments to reach goals.
Unlike vision-and-language navigation (VLN)~\cite{r2r,vlnce,rxr}, GN is independent of starting points and requires agents to explore and localize goals without step-by-step guidance.
Existing tasks specify goals in various forms, such as point coordinates (PointNav)~\cite{anderson2018evaluation, chaplot2020learning, partsey2022mapping}, object categories (ObjectNav)~\cite{habitatchallenge, chaplot2020object, batra2020objectnav, zhang20233d, Chen2023HowToNotTrainYourDragon,ovon,wani2020multion}, reference images (ImageNav)~\cite{zhu2017target,krantz2022instance,krantz2023navigating,kim2023topological}, or VLM-generated instance descriptions~\cite{goatbench,PSL}.

GN methods typically follow two paradigms: end-to-end reinforcement learning and modular architectures.
End-to-end reinforcement learning methods~\cite{Wijmans2020DDPPO,Ye2021AuxiliaryObjectNav,ramrakhya2023pirlnav,Zeng2024PoliFormer} directly map inputs to low-level actions but often exhibit limited long-horizon reasoning and interpretability.
Modular methods~\cite{Yu2023FrontierSemanticExploration,yokoyama2024vlfm} decompose navigation into components and build explicit scene and object representations, such as scene graphs~\cite{yin2025unigoal,feng2025hyperrectangle,feng2025history} or top-down maps~\cite{Ramakrishnan2022PONI,Georgakis2022ActiveSemanticNav}, but mainly rely on depth sensors and object detection/segmentation models~\cite{ravi2024sam2,ravos}.
Recent advances in VLMs~\cite{clip,hurst2024gpt4o,bai2025qwen2vl} enable zero-shot, open-vocabulary navigation~\cite{Majumdar2022ZSON,gadre2023cows,chen2023open,zhou2023esc,long2024instructnav,yin2025unigoal} through multimodal alignment and broad open-world knowledge.
Despite these advances, prior work has largely overlooked language-specified goals across multiple semantic levels, especially in mixed-level multi-goal episodes spanning scene, room, region, and instance levels.
In addition, strong RGB-only baselines for multi-goal navigation remain underexplored, particularly those operating without depth, explicit 3D representations, or object detection.

\noindent\textbf{Language-Conditioned Goal Navigation Benchmarks.}
Early LGN benchmarks~\cite{habitatchallenge,batra2020objectnav} use real-world scans~\cite{hm3d,hm3dsem,mp3d} to evaluate generalization to unseen environments, but cover only 6--21 object categories.
To improve scene diversity, ProcTHOR~\cite{ProcTHOR} generates floor plans populated with 3D assets, and OVMM~\cite{homerobot} curates interactive synthetic scenes. Generative methods also synthesize 3D objects and scenes~\cite{wu2023sketch,wu2024external,wu2026hierascaffold}. However, synthetic environments can introduce synthetic-to-real generalization gaps~\cite{hssd}.
To expand object diversity, HM3D-OVON~\cite{ovon} presents an open-vocabulary benchmark and LHPR-VLN~\cite{lhvln} incorporates heuristically inferred room types. Nonetheless, these benchmarks emphasize object detection with limited higher-level semantic reasoning.

To support instance-level goals, GOAT-Bench~\cite{goatbench} and PSL~\cite{PSL} use VLMs to generate descriptions from object views. However, VLMs often fail to capture distinctive cues~\cite{li2023evaluating} and show limited 3D spatial reasoning~\cite{cai2025spatialbot,chen2024spatialvlm}, leading to ambiguous or inaccurate instructions.
Language-guided target localization~\cite{refhuman,sgmg,htr} and object pose estimation~\cite{liu2026sinref6d,liu2025diff9d,liu2026survey} have also been studied without active exploration.
Static 3D visual grounding datasets, such as ScanRefer~\cite{chen2020scanrefer} and Nr3D/Sr3D~\cite{achlioptas2020referit3d}, study fine-grained language grounding in fully observed scenes, but do not require active exploration or navigation under partial observability.
Current navigation benchmarks therefore leave two key gaps: noisy instance-level annotations that hinder reliable evaluation and limited coverage of language-specified goals across semantic levels.
Our benchmark addresses both by providing human-verified contrastive annotations and supporting single-goal and mixed-level multi-goal tasks spanning scene, room, region, and instance levels.

\begin{figure*}[t!]
\centering
\includegraphics[width=1.0\linewidth]{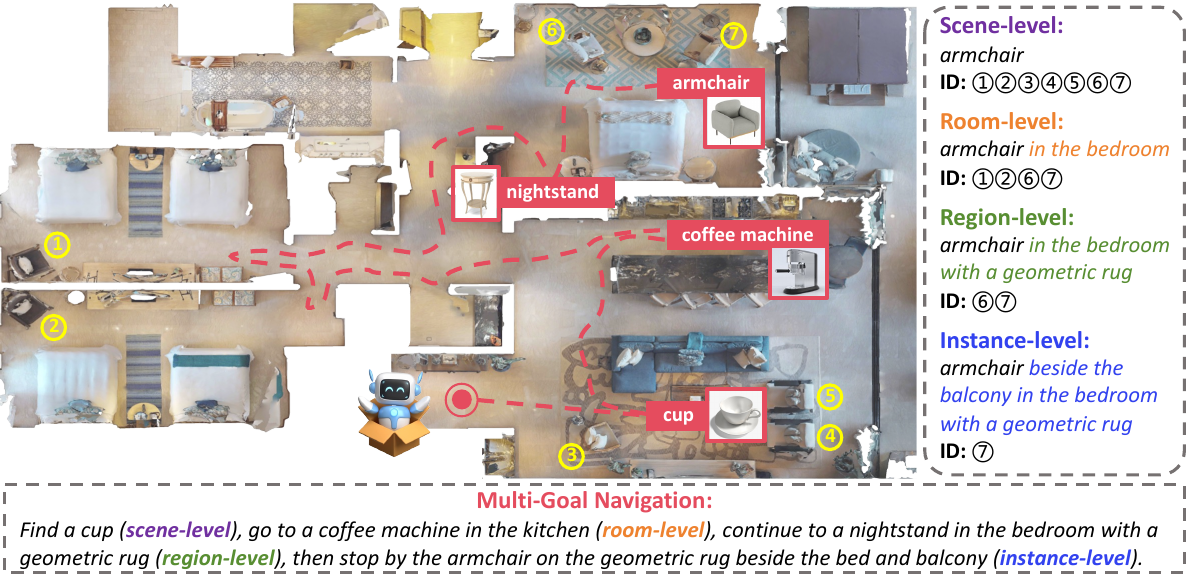}
\caption{
\ourtask\ requires agents to navigate to language-specified targets at four levels: scene, room, region, and instance.
\ourbenchmark\ enables evaluation with mixed-level multi-goal episodes and single-goal tasks.
Descriptions cover intrinsic attributes, spatial context, and open-world knowledge.
}
\label{fig:task_pipeline}
\end{figure*}

\section{Task and Benchmark}

\subsection{\ourtask: Hierarchical Open-Vocabulary Goal Navigation}
\ourtask\ (Figure~\ref{fig:task_pipeline}) requires an agent, randomly initialized in a 3D environment, to complete either multi-goal episodes or single-goal tasks. Evaluation focuses on unseen environments.
Unlike prior benchmarks~\cite{ovon,goatbench,PSL}, \ourtask\ specifies goals in natural language across four semantic levels:

\begin{itemize}[leftmargin=1.2em, itemsep=0.3pt, topsep=1pt]
\item \textbf{Scene-level}: any object of the target category in the scene 
(\eg, ``\textit{armchair}'').
\item \textbf{Room-level}: an object of the target category in a specified room type
(\eg, ``\textit{armchair in the bedroom}'').
\item \textbf{Region-level}: an object of the target category in a specific room instance, distinguished from same-type rooms by contextual cues
(\eg, ``\textit{armchair in the bedroom with a geometric rug}'').
\item \textbf{Instance-level}: a unique object instance identified by discriminative attributes or contextual relations
(\eg, ``\textit{square coffee table}'', ``\textit{armchair beside the bed and balcony}'').
\end{itemize}

At each time step $t$, the agent receives an RGB observation $I_t$, relative odometry $P_t=(\Delta x,\Delta y,\Delta \theta)$, and optional depth $D_t$.
Following standard protocols~\cite{ovon,goatbench}, the action space includes \texttt{MOVE\_FORWARD} (0.25m), \texttt{TURN\_LEFT} or \texttt{TURN\_RIGHT} ($30^\circ$), and \texttt{STOP},
with success defined as executing \texttt{STOP} within 1m of any valid target within 500 steps per goal.
The agent follows Stretch robot specifications~\cite{stretch,ovon}: height 1.41m, base radius 0.17m, and an RGB-D camera mounted at 1.31m.

\subsection{\ourbenchmark~Benchmark Statistics}

Built on real-world HM3D scans~\cite{hm3d}, \ourbenchmark\ covers all 36 HM3D-Sem validation scenes~\cite{hm3dsem} and provides human-verified annotations and tasks across four semantic levels.
Table~\ref{tab:benchmark_stat} highlights its broader goal coverage, greater task diversity, higher annotation quality, and larger task scale.

\begin{figure*}[t!]
\centering
\includegraphics[width=1.0\linewidth]{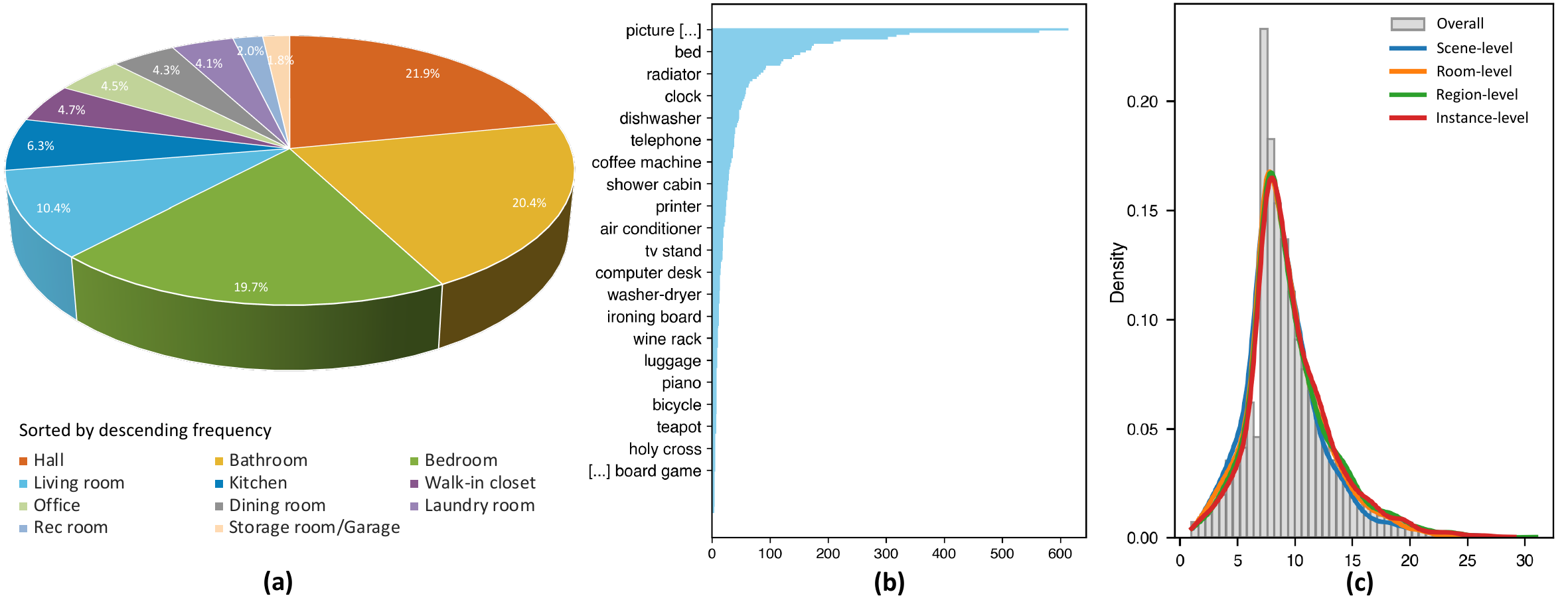}
\caption{
    \ourbenchmark\ distributions. (a) Region label frequency. (b) Instances per object category, with sampled labels. (c) Ground-truth geodesic distances of navigation tasks across four semantic levels.
}
\label{fig:total_distribution}
\end{figure*}

\noindent\textbf{Region Annotations.}
\ourbenchmark\ provides human-verified region annotations across 12 room categories and 926 discriminative region descriptions. These annotations enable room- and region-level goal navigation tasks.
As shown in Figure~\ref{fig:total_distribution}(a), common indoor spaces (\eg, halls, bathrooms, bedrooms) dominate, while recreation rooms, storage rooms, and garages are less frequent.

\noindent\textbf{Object Annotations.}
\ourbenchmark\ covers 414 object categories, with 349 used for navigation tasks: \textbf{2.93$\times$} the number in GOAT-Bench's evaluation split (119) and \textbf{1.34$\times$} that in full GOAT-Bench (260)~\cite{goatbench}.
Compared with~\cite{ovon,goatbench}, \ourbenchmark\ includes more small-object categories (\eg, knife) and retains small but visible targets for more realistic evaluation. Figure~\ref{fig:total_distribution}(b) shows the sorted per-category instance counts.
To enable reliable evaluation, \ourbenchmark\ provides human-verified discriminative descriptions, whereas GOAT-Bench~\cite{goatbench} uses VLM-generated descriptions that often contain semantic errors or lack discrimination (Figure~\ref{fig:goat_error_distribution}).
Concise descriptions average \textbf{5.3} words, providing minimal yet sufficient cues for natural goal specification.

\noindent\textbf{Task Granularity.}
\ourbenchmark\ supports mixed-level multi-goal episodes and single-goal tasks across four semantic levels: scene, room, region, and instance.
Figure~\ref{fig:total_distribution}(c) shows consistent geodesic distance distributions across levels, mostly 5-15\,m, avoiding path-length bias.
Together with human-verified annotations, \ourbenchmark\ enables rigorous evaluation of language-driven embodied navigation.

\noindent\textbf{Data Availability.}
The \ourbenchmark\ dataset and \ourmodel\ code are available at \url{https://github.com/bo-miao/LangMap}, with documentation, licenses, and responsible AI information in the Appendix.
HM3D and HM3D-Sem assets are not redistributed and must be obtained from the official sources under their original licenses.

\begin{table*}[t!]
\setlength{\tabcolsep}{1.4pt}
\centering
\small
\caption{
Statistics of LGN evaluation benchmarks.
Cat/Desc/Words: number of categories, number of discriminative descriptions, and average word count (concise/detailed).
Small Obj: percentage of objects with mean frame coverage below 3.3\% across look-down, forward, and look-up views.
Blanks and $\times$ indicate unsupported settings.
$^{\dagger}$: VLM-generated descriptions (GOAT-Bench) and inferred room types (LHPR-VLN).
\ourbenchmark\ provides human-verified multi-granularity annotations.
}\label{tab:benchmark_stat}
\begin{tabular}{l | c c c c | c c c | c c c | c | c}
\toprule[1.5pt]
\multirow{2}{*}{Eval Benchmark} & \multicolumn{4}{c |}{\textbf{Goal Granularity}} & \multicolumn{3}{c |}{Region Annotation} & \multicolumn{3}{c |}{Object Annotation} & \multirow{2}{*}{Small Obj} & \multirow{2}{*}{Tasks} \\ 
& Scene & Room & Region & Instance & Cat & Desc & Words & Cat & Desc & Words & & \\
\midrule 
RoboTHOR\cite{robothor} & \checkmark & && & $\times$ & $\times$ & $\times$ & 12 & $\times$ & $\times$ & -  & - \\ 
ObjNav-MP3D\cite{batra2020objectnav} & \checkmark & && & $\times$ & $\times$ & $\times$ & 21 & $\times$ & $\times$ & - & - \\ 
ObjNav-HM3D\cite{habitatchallenge} & \checkmark & && & $\times$ & $\times$ & $\times$ & 6 & $\times$ & $\times$  & - & - \\ 
\midrule
HM3D-OVON\cite{ovon} & \checkmark & && & $\times$ & $\times$ & $\times$ & 178 & $\times$ & $\times$ & 4.2\% & 9000 \\
LHPR-VLN\cite{lhvln} & & \checkmark$^{\dagger}$ & &  & 10 & $\times$ & $\times$ & - & $\times$ & $\times$ & - & 960 \\ 
GOAT-Bench\cite{goatbench} & \checkmark & & & \checkmark$^{\dagger}$ & $\times$ & $\times$ & $\times$ & 119 & 1506$^{\dagger}$ & 21.0 & 4.9\% & 7951 \\
\textbf{\ourbenchmark\ (Ours)} & \checkmark & \checkmark & \checkmark & \checkmark & \textbf{12} & \textbf{926} & \textbf{5.7}/21.0 & \textbf{349} & \textbf{7510} & \textbf{5.3}/15.9 & \textbf{22.2\%} & \textbf{18479}  \\ 
\bottomrule[1.5pt]
\end{tabular}
\end{table*}

\subsection{Contrastive Annotation}
\begin{figure*}[t!]
\centering
\includegraphics[width=1.0\linewidth]{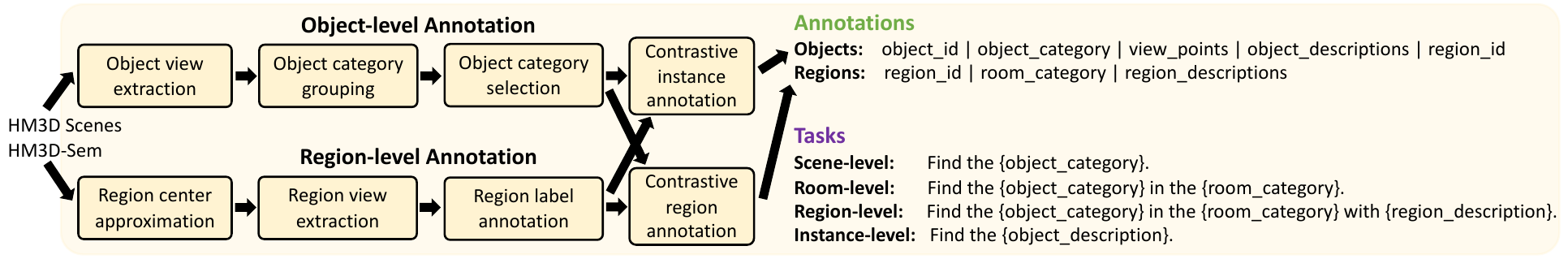}
\caption{Overview of our data processing pipeline.}
\label{fig:data_processing}
\vspace{-1mm}
\end{figure*}

We design a contrastive annotation process to produce discriminative referring expressions for regions and object instances (Figure~\ref{fig:data_processing}).
Starting from HM3D scenes~\cite{hm3d} and HM3D-Sem annotations~\cite{hm3dsem}, we use region--object mappings, object labels, positions, and bounding boxes to guide view extraction.
For each object instance, we select a representative view with maximal visible coverage. For each region, we capture a panorama at a pseudo-center defined as the midpoint of the bounding box enclosing its objects.
Together, these views and metadata provide the context for contrastive annotation.
The resulting annotations are used to generate single-goal tasks and mixed-level multi-goal episodes across four semantic levels. View extraction details are provided in Appendix.

\begin{figure*}[t!]
\centering
\includegraphics[width=1.0\linewidth]{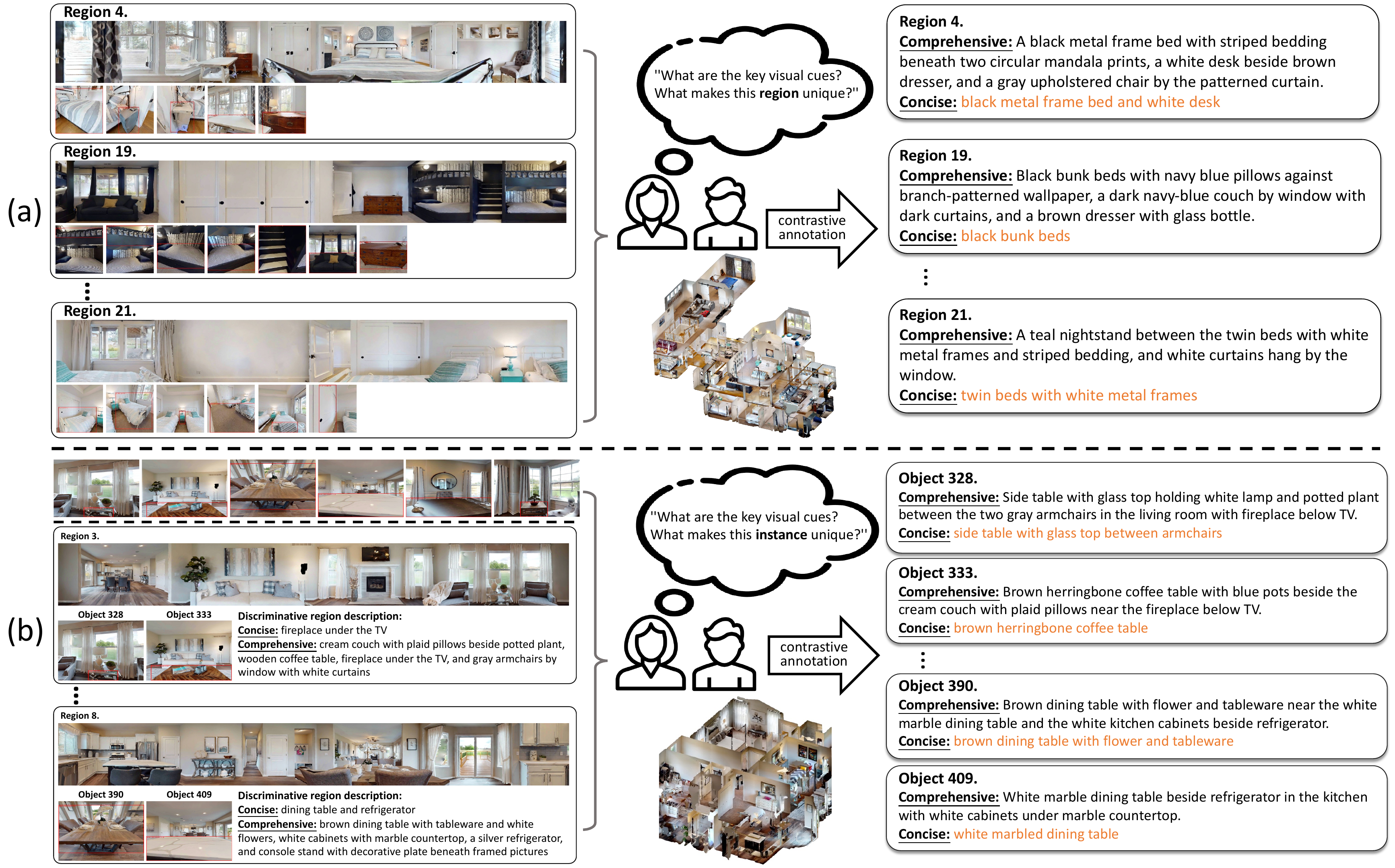}
\caption{Examples of the contrastive annotation process.
(a) Region annotation: annotators compare same-category regions within a scene.
(b) Instance annotation: annotators compare same-category or semantically related instances using object views and verified region context.
All annotations are cross-checked for correctness and discriminability.}
\label{fig:benchmark_curation_total}
\end{figure*}

\noindent\textbf{Contrastive Region Annotation.}
Given region panoramas and their contained object views, annotators first label room categories; regions spanning multiple types receive all applicable labels (\eg, a living room connected to a kitchen).
They then compare same-category regions within each scene to compose concise and detailed descriptions that uniquely identify each region (Figure~\ref{fig:benchmark_curation_total}(a)).
Concise descriptions capture the most distinctive cue, and detailed descriptions extend them with visual attributes, objects, and spatial context.

\noindent\textbf{Contrastive Instance Annotation.}
To address fine-grained object label ambiguity, we cluster semantically similar object categories using SentenceBERT~\cite{reimers2019sentence} and refine them into a hierarchy, where fine-grained classes are grouped under base categories (\eg, \textit{coffee table} under \textit{table}).
For each base category, annotators compare all instances across its fine-grained categories within a scene and write concise and detailed descriptions that distinguish each target from same-category distractors (Figure~\ref{fig:benchmark_curation_total}(b)).
For dense object categories such as cabinets, verified region context narrows comparisons to within-region candidates, while region cues in the descriptions preserve scene-level uniqueness.
Instance descriptions cover intrinsic attributes (\eg, color, material, pattern, shape, and size), spatial context (\eg, relative position), and open-world semantics (\eg, an Eiffel Tower photo).

\subsection{Annotation Protocol and Quality Control}
All descriptions are written by seven researchers (three PhD holders and four research students) following shared guidelines and calibration examples, with VLM-generated descriptions~\cite{hurst2024gpt4o} used only as optional references.
After initial annotation, scenes are shuffled and reassigned for cross-checking of correctness, label consistency, and discriminability. About 23\% of first-round annotations are revised, primarily for insufficient discriminability rather than factual errors. Two lead annotators resolve disagreements.
Annotation and review take approximately 630 person-hours in total.
Further details are provided in the Appendix, and automated and human evaluations are reported in Sec.~\ref{sec:annotation_quality}.

\subsection{Navigation Task Generation}
A single-goal task comprises a scene, a sampled start pose, and an instruction specifying a goal at one of four semantic levels with one or more valid targets; a multi-goal episode chains such goals to be completed in order.
For scene-level tasks, we enumerate all object categories to prevent duplication.
For room-level tasks, each object category is paired with every room type containing it, and targets are restricted to that category and room type using HM3D-Sem region--object mappings~\cite{hm3dsem} and our human-labeled room categories.
For region-level tasks, this target set is further narrowed by a discriminative region description, and instance-level tasks use discriminative instance descriptions as instructions.
For each single-goal task, we randomly sample a start pose such that at least one valid target lies on the same floor~\cite{ovon,goatbench} and the geodesic distance to the nearest valid target is 5--30m, with the lower bound relaxed to 1m if no such pose exists. This yields about 15K single-goal tasks.
For multi-goal episodes, we uniformly sample five same-floor tasks spanning multiple semantic levels with non-overlapping targets, yielding 720 episodes (3.6K tasks), with no pose reset between goals.

\section{Method}
\label{sec:method}
\vspace{-1mm}

\begin{figure*}[h]
\centering
\includegraphics[width=1.0\linewidth]{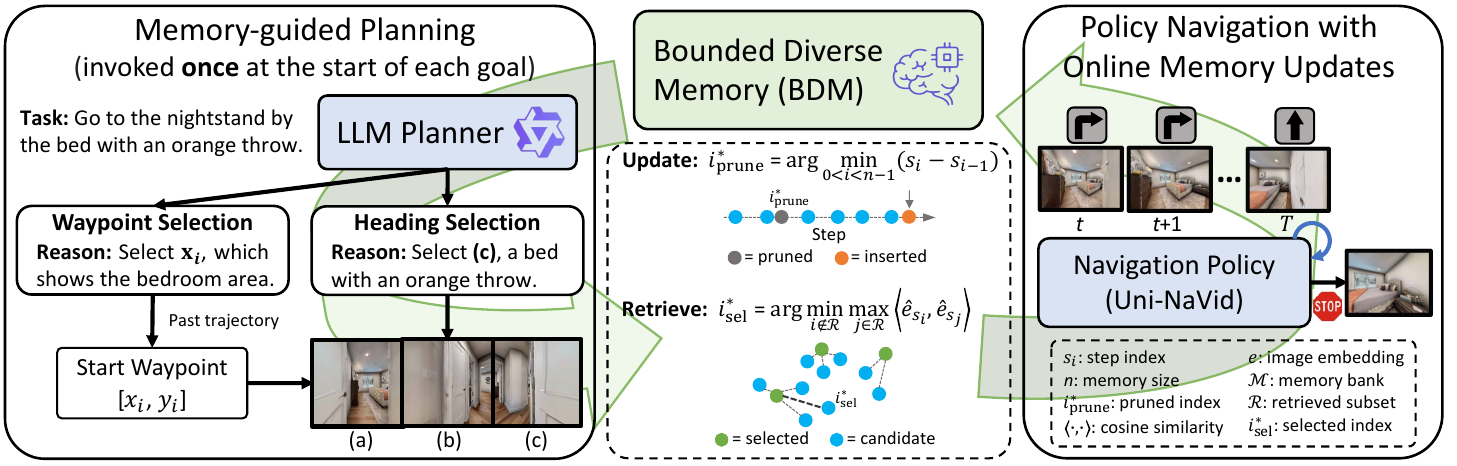}
\caption{
Overview of~\ourmodel, a decoupled RGB-only framework using RGB and language \textit{without} depth, 3D representations, or object masks.
Bounded Diverse Memory maintains a bounded history with near-uniform temporal coverage and provides diverse context for a VLM planner, which selects an initial waypoint and heading per goal to prime reactive policy for multi-goal navigation.
}
\label{fig:method}
\end{figure*}

\subsection{Decoupled RGB-based Framework}

As shown in Figure~\ref{fig:method}, \ourmodel\ decouples long-horizon planning from reactive navigation in two stages.
In memory-guided planning, \ourmemory\ $\mathcal{M}$ retrieves semantically diverse context for a VLM planner to select a goal-relevant initial waypoint and heading.
In policy navigation, a goal-conditioned reactive policy maps RGB observations to actions and updates $\mathcal{M}$ online for subsequent goals.
We adopt Qwen2.5-VL-7B-Instruct~\cite{bai2025qwen2vl} as the planner and Uni-NaVid~\cite{uninavid} as the policy.
Unlike methods relying on depth, 3D scene representations, or object masks, \ourmodel\ maintains a bounded memory of semantically diverse RGB snapshots for multi-goal navigation.

\subsection{\ourmemory}
\label{subsec:memory}

Our memory maintains at most $N_{\max}$ RGB snapshots with near-uniform temporal coverage and semantic diversity for long-horizon reasoning.
It has two components: global-uniform update (GUU) and semantic-diverse retrieval (SDR). The pseudocode is provided in the Appendix.

\subsubsection{Global-Uniform Update}
In~\ourtask, multi-goal episodes can span up to 2,500 steps, making full-trajectory storage computationally prohibitive and redundant given the limited context windows of VLM planners.
To maintain a compact yet globally representative history, we perform global-uniform update to ensure near-uniform temporal coverage under a fixed budget $N_{\max}$.
At each step $t$, we store $\mathbf{x}_t=(s_t,\mathbf{I}_t,\mathbf{e}_t,\mathbf{p}_t)$, containing the step index, RGB observation, image embedding, and agent position.
When the buffer size exceeds $N_{\max}$, gap-aligned pruning removes the entry with minimum local temporal gap:
\begin{equation}
i^*_{\text{prune}}=\arg\min_{0<i<n-1}(s_i-s_{i-1}).
\end{equation}
\noindent This reduces temporal redundancy and maintains near-uniform coverage under a fixed capacity.

\noindent\textbf{Temporal Coverage Error Bound.}
Let $\mathcal{S}=\{s_i\}_{i=0}^{n-1}$ be the retained step indices in increasing order, 
with $s_0=0$, $s_{n-1}=T$, and $n\le N_{\max}$.
For any $t\in[0,T]$, the temporal approximation error is $\epsilon(t)\triangleq \min_{0\le i< n}|t-s_i|$.
Let $g_i=s_i-s_{i-1}$ and $g_{\max}=\max_i g_i$. The \textit{worst-case error} then satisfies $\epsilon_{\max}\triangleq\max_{t\in[0,T]}\epsilon(t)\le g_{\max}/2$, and if gaps are near-uniform with spacing $\Delta\approx T/(n-1)$, the \textit{average per-step error} is $\bar\epsilon\approx \Delta/4$.
Since the worst-case error is governed by the largest temporal gap, near-uniform spacing is desirable under a fixed budget, motivating our gap-aligned pruning.
With $N_{\max}=50$ and a typical $T\approx700$ ($\Delta\approx14$), near-uniform spacing gives $\epsilon_{\max}\approx7$ and $\bar{\epsilon}\approx3.5$ steps while compressing memory by \textbf{92.9\%}.

\subsubsection{Semantic-Diverse Retrieval}
To fit the VLM planner's context window, we retrieve up to $K$ memory states ($K{=}10$ in this work) to promote semantic coverage.
Given memory $\mathcal{M}$, we anchor the most recent state $\mathbf{x}_{s_{n-1}}$ since it incurs no movement cost.
We then iteratively select the remaining $K-1$ states using a greedy max--min strategy for global semantic diversity:
\begin{equation}
i^*_{\text{select}} = \arg\min_{i \notin \mathcal{R}}
\max_{j \in \mathcal{R}}
\langle \hat{\mathbf{e}}_{s_i}, \hat{\mathbf{e}}_{s_j} \rangle,
\end{equation}
where $\mathcal{R}$ denotes the selected indices, $\hat{\mathbf{e}}_{s_i}$ the normalized image embedding, and $\langle\cdot,\cdot\rangle$ cosine similarity.
After the planner chooses a waypoint from the retrieved subset, the agent replays the compressed recorded trajectory to reach it without querying the simulator pathfinder.

\section{Experiments}
\label{sec:Experiments}

\subsection{Metrics}
\vspace{-1mm}
We report Success Rate (SR) and Success weighted by Path Length (SPL)~\citep{ovon,goatbench,anderson2018evaluation}.
SR and SPL measure per-goal performance but not sequential reliability: a single failed goal prevents completion of the entire sequence, making partial success insufficient for deployment (\eg, finding a cup but failing to reach the coffee machine).
We therefore define Sequence Success Rate at $k$ (SeqSR@$k$) as the fraction of episodes in which the first $k$ goals are successfully completed in order:
{\small
\begin{equation}
\mathrm{SeqSR}\text{@}k = \frac{1}{N_\mathrm{ep}} \sum_{i=1}^{N_\mathrm{ep}}
\mathbb{I}\!\left(\sum_{j=1}^{k} S_{i,j} = k\right),
\end{equation}
}
where $N_{\mathrm{ep}}$ is the number of episodes, $S_{i,j}\in\{0,1\}$ denotes success on goal $j$ of episode $i$, and each episode contains $G{=}5$ goals ($k\le G$).

\begin{table}[t]
\centering
\small
\setlength{\tabcolsep}{1.5pt}
\caption{
Evaluation on~\ourbenchmark\ with \textbf{concise descriptions}.
-3B/-7B denote planner sizes.
$^{\dagger}$Reference only: 3D-Mem's original success criterion (within 1m of a target viewpoint, allowing up to 2m from the goal).
Local Nav.: simulator pathfinder (Shortest) or learned policy (Learned).
}
\label{tab:mainresults}
\begin{tabular}{l| c c c c | ccc | cc | c}
\toprule[1.5pt]
\multirow{2}{*}{Method} & 
\multirow{2}{*}{Observation} &
\multirow{2}{*}{Mask} &
\multirow{2}{*}{3D Map} &
\multirow{2}{*}{Local Nav.} &
\multicolumn{3}{c|}{\textbf{Multi-Goal}} & 
\multicolumn{2}{c|}{\textbf{Single-Goal}} &
\multirow{2}{*}{\shortstack{Latency\\(s/goal)}} \\
\cmidrule(lr){6-8} \cmidrule(lr){9-10}
& & & & &
SR$\uparrow$ & SeqSR@2$\uparrow$ & SPL$\uparrow$
& SR$\uparrow$ & SPL$\uparrow$ & \\
\midrule
3D-Mem-3B~\cite{3dmem} & RGB-D (pano.) &\checkmark &\checkmark & Shortest & 13.5&1.1&6.4 & 8.5&1.9  & 147.2 \\  
3D-Mem-7B~\cite{3dmem} & RGB-D (pano.) &\checkmark &\checkmark & Shortest & 30.1&5.1&17.3 & 13.7&6.2 & 95.6 \\ 
\faded{3D-Mem-3B$^{\dagger}$}~\cite{3dmem} & \faded{RGB-D (pano.)} & \faded{\checkmark} & \faded{\checkmark} & \faded{Shortest} & \faded{20.4} & \faded{2.2} & \faded{11.3} & \faded{15.3} & \faded{2.8} & \faded{147.2} \\ 
\faded{3D-Mem-7B$^{\dagger}$}~\cite{3dmem} & \faded{RGB-D (pano.)} & \faded{\checkmark} & \faded{\checkmark} & \faded{Shortest} & \faded{36.8} & \faded{10.0} & \faded{21.2} & \faded{21.2} & \faded{8.7} & \faded{95.6} \\ 
MTU3D~\cite{mtu3d} & RGB-D (pano.) &\checkmark &\checkmark & Shortest & 41.2&11.0&\cellcolor{teal!15}\textbf{24.3}  & 29.9&15.3 & 52.1 \\  
\midrule
PSL~\cite{PSL} & RGB (single) & $\times$ & $\times$ & Learned & 14.6&3.1&4.0 & 13.8&4.7 & - \\ 
SenseAct-M~\cite{goatbench} & RGB (single) & $\times$ & $\times$ & Learned & 15.5&1.0&8.4  &  8.7&4.6  & - \\ 
Uni-NaVid~\cite{uninavid} & RGB (single) & $\times$ & $\times$ & Learned & 34.4&10.4&15.0  &  30.3&15.3 & 19.0 \\ 

\textbf{\ourmodel-3B} & RGB (single) & $\times$ & $\times$ & Learned & 41.5&14.4&17.5 & 31.3&15.2 & 19.4 \\  
\textbf{\ourmodel-7B} & RGB (single) & $\times$ & $\times$ & Learned & \cellcolor{teal!15}\textbf{42.8}& \cellcolor{teal!15}\textbf{14.6}&18.1 
& \cellcolor{teal!15}\textbf{31.7}&\cellcolor{teal!15}\textbf{15.4} & 19.7 \\ 

\bottomrule[1.5pt]
\end{tabular}
\vspace{-2mm}
\end{table}

\begin{table}[t]
\centering
\small
\setlength{\tabcolsep}{2.7pt}
\caption{Single-goal results on \ourbenchmark. Descriptions are concise unless marked `-D' (detailed).}
\label{tab:single_goal}
\begin{tabular}{l|l|cc|cc|cc|cc|cc|cc}
\toprule[1.5pt]
Config & Method & \multicolumn{2}{c|}{\textbf{Scene}} & \multicolumn{2}{c|}{\textbf{Room}} & \multicolumn{2}{c|}{\textbf{Region}} & \multicolumn{2}{c|}{\textbf{Region-D}} & \multicolumn{2}{c|}{\textbf{Instance}} & \multicolumn{2}{c}{\textbf{Instance-D}} \\
\cmidrule(lr){3-4} \cmidrule(lr){5-6} \cmidrule(lr){7-8} \cmidrule(lr){9-10} \cmidrule(lr){11-12} \cmidrule(lr){13-14}
& & SR$\uparrow$ & SPL$\uparrow$ & SR$\uparrow$ & SPL$\uparrow$ & SR$\uparrow$ & SPL$\uparrow$ & SR$\uparrow$ & SPL$\uparrow$ & SR$\uparrow$ & SPL$\uparrow$ & SR$\uparrow$ & SPL$\uparrow$ \\
\midrule
\multirow{3}{*}{RGB-D,Mask}
& 3D-Mem-3B & 11.2&2.5 & 11.8&2.8 & 8.2&1.8 & 6.8&1.4   & 4.7&1.1 & 5.1&1.1 \\ 
& 3D-Mem-7B & 16.4 & 7.2 & 12.6 & 6.0 & 14.0 & 6.1 & 15.1 & 5.6 & 13.0 & 5.8 & 16.3 & 7.0\\ 
& MTU3D & 31.4&15.7 & 32.5&16.2 &  \cellcolor{teal!15}33.6&\cellcolor{teal!15}16.8 & \cellcolor{teal!15}33.7&16.2 & 23.8&12.9 & 28.7&15.2 \\ 
\midrule
\multirow{5}{*}{RGB}
& PSL & 16.5 & 5.3 & 16.4 & 5.9 & 12.8 & 4.3 & 13.1 & 4.4 & 11.2 & 3.8 & 10.3 & 3.6 \\ 
& SenseAct-M & 10.2 & 5.6 & 8.3 & 4.4 & 8.5 & 4.3 & 9.5 & 4.8 & 7.7 & 4.0 & 7.4 & 4.1 \\ 
& Uni-NaVid & 33.8 & 16.2 & 33.2 & 16.5 & 30.1 & 15.5 & 31.7 &\cellcolor{teal!15}17.4 & 26.2 & \cellcolor{teal!15}13.8 & 28.8&15.9 \\ 
& \textbf{\ourmodel-3B} & 35.1 & 16.8 & \cellcolor{teal!15}36.0 & \cellcolor{teal!15}16.6 & 30.7 & 15.3 & 32.7&16.7 & 26.3 & 13.1 & 29.4&15.6 \\  
& \textbf{\ourmodel-7B} & \cellcolor{teal!15}35.8&\cellcolor{teal!15}17.2 & 34.7&16.4 & 31.1&15.3 & 33.2&17.2 & \cellcolor{teal!15}27.8&13.7  & \cellcolor{teal!15}29.6&\cellcolor{teal!15}15.9\\  
\bottomrule[1.5pt]
\end{tabular}
\vspace{-1mm}
\end{table}

\subsection{Main Results}
Tables~\ref{tab:mainresults} and~\ref{tab:single_goal} compare recent LGN methods on \ourbenchmark, distinguishing their sensing, mapping, and local navigation resources.
PSL and SenseAct-NN Monolithic lag well behind the other RGB-only methods. PSL is limited by CLIP’s weak compositional reasoning~\citep{clip,kamath2023text} and closed-set training data, and noisy GOAT-Bench supervision may affect SenseAct-NN.
3D-Mem~\cite{3dmem} combines VLM reasoning with an occupancy map and 3D object features. For reproducibility, we evaluate it zero-shot using the same off-the-shelf Qwen2.5-VL-3B/7B models~\cite{bai2025qwen2vl} as \ourmodel\ rather than closed-source models. Its multi-goal SR rises with planner size (30.1\% for 7B vs.\ 13.5\% for 3B), but latency is about $5\times$ that of~\ourmodel\ (95.6 vs.\ 19.7\,s/goal on an RTX 4090).
MTU3D and Uni-NaVid are trained on million-scale data. MTU3D uses panoramic RGB-D for map-based reasoning, whereas Uni-NaVid is a single-view RGB reactive policy. Uni-NaVid, without depth or explicit high-level planning, attains slightly higher single-goal SR but trails MTU3D by 6.8 points in multi-goal SR.

Our RGB-only \ourmodel\ achieves the highest multi-goal and single-goal SR (\textbf{42.8\%} and \textbf{31.7\%}).
Under identical RGB sensing, it improves multi-goal SR over Uni-NaVid by \textbf{8.4} points, validating memory-guided decoupled design.
It also achieves SR comparable to panoramic MTU3D, which uses depth, object masks, 3D maps, and simulator shortest-path navigation, though with lower multi-goal SPL. Under matched horizontal fields of view, \ourmodel-7B outperforms MTU3D by 4.8 points in multi-goal SR.
\ourmodel\ is also less sensitive to planner size than 3D-Mem.
Across methods, performance is generally higher at coarser levels than at the instance level, where finer disambiguation is required, and detailed descriptions often improve region- and instance-level results.
All methods achieve consistent gains in multi-goal navigation, likely from implicit temporal encoding or explicit memory, but low SeqSR indicates that completing goal sequences remains challenging.
Full SeqSR@1--5 results are reported in the Appendix.

\subsection{Annotation Quality Analysis}
\label{sec:annotation_quality}
\vspace{-1mm}
To evaluate annotation discriminability against GOAT-Bench~\cite{goatbench}, we extract all overlapping annotated instances and perform one-to-many text-to-view matching using GPT-4o~\cite{hurst2024gpt4o} and Qwen3-VL-235B-A22B~\cite{Qwen3-VL}.
Since region annotations are unique to~\ourbenchmark, we focus this comparison on instance-level descriptions.
Table~\ref{tab:ablation_bench_compare} shows that \ourbenchmark\ achieves about \textbf{80\%} matching accuracy, compared with 57\% for GOAT-Bench, while using about one quarter as many words, and is non-inferior in 94\% of cases.
We further conduct an independent human audit, where ten auditors match a stratified sample of descriptions to targets among 5--10 same-scene distractors, yielding 92.5\% unique-and-correct matches across 200 judgments. It evaluates description discriminability and correctness rather than inter-annotator agreement.
These results show that our annotations are more concise, discriminative, and reliable.
Extensive visual comparisons are provided in the Appendix.

\subsection{Failure Diagnostics Across Goal Levels}
\label{sec:failure}
\vspace{-1mm}
Table~\ref{tab:failure_perlevel} reports OSR (success-radius entry) over all subgoals and the shares of exploration (target neither perceived nor reached), hierarchical disambiguation (stopped at the wrong room, region, or instance), and termination/control (target found but no successful stop) among failed subgoals under first-hit attribution.
From scene- to instance-level goals, \ourmodel's OSR drops from 69.9\% to 36.7\%, and the dominant failure category shifts from termination/control (75.7\%) to exploration (48.9\%), with hierarchical errors accounting for 19.7\% of instance-level failures.

\begin{table*}[t!]
\centering
\begin{minipage}{.55\textwidth}
\centering
\small
\setlength{\tabcolsep}{2pt}
\caption{
Annotation discriminability under one-to-many text-to-view matching.
\textit{Exclusive Win:} instances correctly matched only with the respective benchmark.
}
\begin{tabular}{l | l|c|c|c}
\toprule[1.5pt]
Evaluator & Benchmark & Words & Accuracy & Excl. Win ($\uparrow$) \\
\midrule
\multirow{2}{*}{GPT-4o}
& GOAT-Bench & 21.1 & 58.4\% & 6.2\% \\
& \textbf{\ourbenchmark} & \cellcolor{teal!15}\textbf{5.2} & \cellcolor{teal!15}\textbf{81.3\%} & \cellcolor{teal!15}\textbf{29.1\%} \\ 
\midrule
\multirow{2}{*}{Qwen3-VL}
& GOAT-Bench & 21.1 & 55.9\% & 5.3\% \\
& \textbf{\ourbenchmark} & \cellcolor{teal!15}\textbf{5.2} & \cellcolor{teal!15}\textbf{79.7\%} & \cellcolor{teal!15}\textbf{29.1\%} \\ 
\bottomrule[1.5pt]
\end{tabular}
\label{tab:ablation_bench_compare}
\end{minipage}
\hfill
\begin{minipage}{.4\textwidth}

\centering
\small
\setlength{\tabcolsep}{4pt}
\caption{Per-level multi-goal OSR and failure diagnostics using \ourmodel-7B.}
\label{tab:failure_perlevel}
\begin{tabular}{l|c|ccc}
\toprule[1.5pt]
Level & OSR$\uparrow$ & Expl. & Hier. & Term. \\
\midrule
Scene    & 69.9 & 24.3 & --   & 75.7 \\
Room     & 67.2 & 22.1 & 12.1 & 65.8 \\
Region   & 61.7 & 28.1 & 10.8 & 61.1 \\
Instance & 36.7 & 48.9 & 19.7 & 31.4 \\
\midrule
Overall  & 59.4 & 31.9 & 11.7 & 56.4 \\
\bottomrule[1.5pt]
\end{tabular}
\end{minipage}
\vspace{-3mm}
\end{table*}

\begin{table}[t]
\centering
\small
\setlength{\tabcolsep}{2.5pt}
\caption{Single-goal analysis by category frequency, optimal path length, and object size.}
\label{tab:ablation_frequency_distance_size}
\begin{tabular}{l|cc|cc||cc|cc|cc||cc|cc}
\toprule[1.5pt]
\multirow{2}{*}{Method} & \multicolumn{4}{c||}{Category Frequency} & \multicolumn{6}{c||}{Path Length} & \multicolumn{4}{c}{Object Size}  \\
\cmidrule(lr){2-5} \cmidrule(lr){6-11} \cmidrule(lr){12-15} 
& \multicolumn{2}{c|}{Head} & \multicolumn{2}{c||}{Long-tail} & \multicolumn{2}{c|}{Short} & \multicolumn{2}{c|}{Medium} & \multicolumn{2}{c||}{Long} & \multicolumn{2}{c|}{Non-small} & \multicolumn{2}{c}{Small}  \\
\cmidrule(lr){2-3} \cmidrule(lr){4-5} \cmidrule(lr){6-7} \cmidrule(lr){8-9} \cmidrule(lr){10-11} \cmidrule(lr){12-13} \cmidrule(lr){14-15} 
& SR$\uparrow$ & SPL$\uparrow$ & SR$\uparrow$ & SPL$\uparrow$ & SR$\uparrow$ & SPL$\uparrow$ & SR$\uparrow$ & SPL$\uparrow$ & SR$\uparrow$ & SPL$\uparrow$ & SR$\uparrow$ & SPL$\uparrow$ & SR$\uparrow$ & SPL$\uparrow$  \\
\midrule
3D-Mem-7B & 14.4 & 6.6 & 11.1 & 4.6 & 27.3 & 12.8 & 11.6 & 4.9 & 4.4 & 2.0 & 14.7 & 6.7 & 11.0 & 4.6 \\
MTU3D & 32.1 & \cellcolor{teal!15}16.3 & 22.0 & 11.5 & \cellcolor{teal!15}46.3 & \cellcolor{teal!15}21.7 & 28.0 & \cellcolor{teal!15}14.8 & 17.4 & 9.7 & 32.8 & \cellcolor{teal!15}16.7 & 20.5 & 10.0 \\
Uni-NaVid & 31.3 & 15.9 & 26.1 & \cellcolor{teal!15}13.2 & 42.2 & 21.5 & 29.7 & 14.7 & 19.4 & 10.3 & 32.6 & 16.5 & \cellcolor{teal!15}23.0 & \cellcolor{teal!15}10.9 \\
\textbf{\ourmodel-3B} & 32.4 & 15.8 & 27.4 & 13.0 & 43.8 & \cellcolor{teal!15}21.7 & 30.5 & 14.4 & \cellcolor{teal!15}20.7 & \cellcolor{teal!15}10.4 & 34.0 & 16.4 & 22.4 & 10.5 \\
\textbf{\ourmodel-7B} & \cellcolor{teal!15}32.8 & 16.1 & \cellcolor{teal!15}27.5 & 12.7 & 43.6 & \cellcolor{teal!15}21.7 & \cellcolor{teal!15}31.2 & \cellcolor{teal!15}14.8 & \cellcolor{teal!15}20.7 & 10.3 & \cellcolor{teal!15}34.5 & \cellcolor{teal!15}16.7 & 22.5 & 10.4 \\
\bottomrule[1.5pt]
\end{tabular}
\vspace{-1mm}
\end{table}

\vspace{-1mm}
\subsection{Ablation Study}
\vspace{-1mm}

\noindent\textbf{Analysis by Category Frequency, Path Length, and Object Size.}
Table~\ref{tab:ablation_frequency_distance_size} reports the results.
Following the Pareto principle, we split categories into head (top 20\%) and long-tail groups, covering 77\% and 23\% of tasks.
We group tasks by optimal path length into short (bottom 25\%), medium (middle 50\%), and long (top 25\%).
All methods perform worse on long-tail categories, and \ourmodel\ achieves highest SR in both groups.
Performance drops with path length, and small targets are harder across methods, highlighting the challenges posed by distant and low-visibility targets.

\begin{wraptable}{r}{0.37\linewidth}
\centering
\footnotesize
\vspace{-4mm}
\setlength{\tabcolsep}{0.7pt}
\caption{Ablation of \ourmodel.}
\vspace{-1mm}
\begin{tabular}{c c c | c c c c}
\toprule[1.5pt]
\multicolumn{3}{c|}{Components} & \multicolumn{4}{c}{Results}  \\
\midrule
Mem & GUU & SDR & SR & SeqSR & SPL & \#Frames$\downarrow$ \\
\midrule
&& & 34.4 & 10.4 & 15.0 & - \\
\checkmark && & 40.4 & 13.2 & 16.9 & 656\\
\checkmark & \checkmark &  & 40.9 & 13.2 & 17.3 & 50 \\
\checkmark && \checkmark & 42.0 & 14.2 & 17.8 & 629 \\
\checkmark & \checkmark & \checkmark & \cellcolor{teal!15}\textbf{42.8} & \cellcolor{teal!15}\textbf{14.6} & \cellcolor{teal!15}\textbf{18.1} & \cellcolor{teal!15}\textbf{50} \\ 
\bottomrule[1.5pt]
\end{tabular}
\label{tab:ablation_study}
\vspace{-5mm}
\end{wraptable}

\noindent\textbf{\ourmemory.}
Table~\ref{tab:ablation_study} ablates each component for multi-goal navigation.
The reactive baseline achieves 34.4\% SR.
Full-trajectory memory (656 frames on average) boosts SR by 6.0 points but is highly redundant.
GUU compresses memory 13$\times$ (656$\rightarrow$50) without performance loss, and SDR improves SR to 42.0\% by retrieving diverse context.
Combining both achieves the best results: 42.8\% SR and 14.6\% SeqSR@2 with a 13$\times$ memory reduction.

\section{Conclusion}
\label{sec:Conclusion}
\vspace{-2mm}
We introduced \ourtask, a hierarchical open-vocabulary goal navigation task spanning scene, room, region, and instance levels. To support reliable evaluation, we presented \ourbenchmark, a real-world 3D indoor navigation benchmark with human-verified annotations and tasks across all four goal levels.
\ourbenchmark\ offers broader coverage, greater task diversity, and higher annotation quality than prior benchmarks.
We also proposed \ourmodel, an RGB-only baseline that employs \ourmemory\ to prime reactive navigation, achieving top-tier success rates without depth, 3D representations, or object masks.
Systematic evaluations on \ourbenchmark\ show the benefits of memory and richer context, while highlighting challenges in exploration, hierarchical disambiguation, timely stopping, and multi-goal completion, as well as navigation to long-tail, small, and distant targets.

\newpage
\section*{Acknowledgements and Disclosures}
This work was supported by the funding received from the Centre for Augmented Reasoning, an initiative by the Commonwealth of Australia. The authors declare no competing interests.

{
\bibliography{egbib.bib}
}


\appendix
\newpage

\section{Dataset Documentation and Access}

\noindent\textbf{Annotation JSON Files.}
Each file contains the following top-level fields:
\texttt{goals}, \texttt{region\_annotation}, \texttt{episodes\_by\_object\_level}, \texttt{episodes\_by\_room\_level}, \texttt{episodes\_by\_region\_level}, \texttt{episodes\_by\_instance\_level}, and \texttt{episode\_by\_sequence}.
The \texttt{goals} field stores object annotations, including the object category, object identifier, 3D position, success viewpoints, associated region identifier, and concise and detailed discriminative instance descriptions.
The \texttt{region\_annotation} field stores the region identifier, human-labeled region category, and concise and detailed discriminative region descriptions.
Each single-goal episode specifies a randomly sampled initial agent pose, a semantic level, and a set of valid target object identifiers. Level-specific fields further define the target object category, room category, region identifier, or instance identifier.
Multi-goal episodes are stored in \texttt{episode\_by\_sequence}. Each multi-goal episode contains one initial pose and an ordered list of subgoals, each referencing an entry in the corresponding single-goal episode list.

\noindent\textbf{Instruction Construction.}
We store structured task metadata rather than fixed instructions, allowing users to generate equivalent instructions with different wording while preserving the same target set.
By default, scene-, room-, region-, and instance-level instructions use the following templates:
\texttt{Find the \{object\_category\}.},
\texttt{Find the \{object\_category\} in the \{room\_name\}.},
\texttt{Find the \{object\_category\} in the \{region\_category\} that has \{region\_description\}.}, and
\texttt{Find the \{instance\_description\}}.

\noindent\textbf{Licenses.}
Our original contributions to LangMap, including annotations, task definitions, and documentation, are released under CC BY-NC 4.0.
HM3D scene assets~\cite{hm3d} and HM3D-Sem semantic files~\cite{hm3dsem} are not redistributed; users must obtain them from the official Habitat data pages\footnote{\url{https://aihabitat.org/datasets/hm3d/}; \url{https://aihabitat.org/datasets/hm3d-semantics/}.} under their original terms for academic, non-commercial research.
Our license does not supersede applicable upstream terms, including those governing information derived from HM3D or HM3D-Sem.

\noindent\textbf{Responsible AI.}
The Croissant file provides Responsible AI metadata on limitations, biases, personal and sensitive information, use cases, social impact, and provenance.

\section{Limitations and Broader Impacts}

\noindent\textbf{Limitations.}
\ourbenchmark\ uses a human-verified contrastive annotation protocol to provide reliable and discriminative descriptions, addressing ambiguities and semantic errors common in VLM-generated annotations.
It also introduces two limitations.
First, human verification increases annotation cost and may limit scaling to larger scene collections. Automated preprocessing and hierarchical, region-guided comparisons reduce this effort. For new scenes with instance and region metadata, \ourtask\ and the pipeline can be reused.
Second, the descriptions may reflect annotator choices in which attributes are emphasized.
Beyond annotation, \ourbenchmark\ covers static indoor HM3D scenes and does not evaluate navigation in outdoor or dynamic environments.
Future work can combine our pipeline with VLM assistance and targeted human verification to improve scalability while preserving quality, and extend the benchmark to outdoor and dynamic environments.

\noindent\textbf{Broader Impacts.}
\ourbenchmark\ is intended to support reliable evaluation of language-conditioned navigation agents with hierarchical goals. Its human-verified annotations reduce ambiguity and enable analysis of long-tail, small-object, and multi-goal navigation challenges.
However, like other embodied AI benchmarks, it could be misused for indoor surveillance or household profiling.
To reduce these risks, we do not redistribute HM3D/HM3D-Sem assets and require users to follow the source dataset licenses and terms of use.

\section{Additional Annotation Details}

\subsection{Representative Object-View and Region-View Capture}
\begin{figure*}[t!]
\centering
\includegraphics[width=1.0\linewidth]{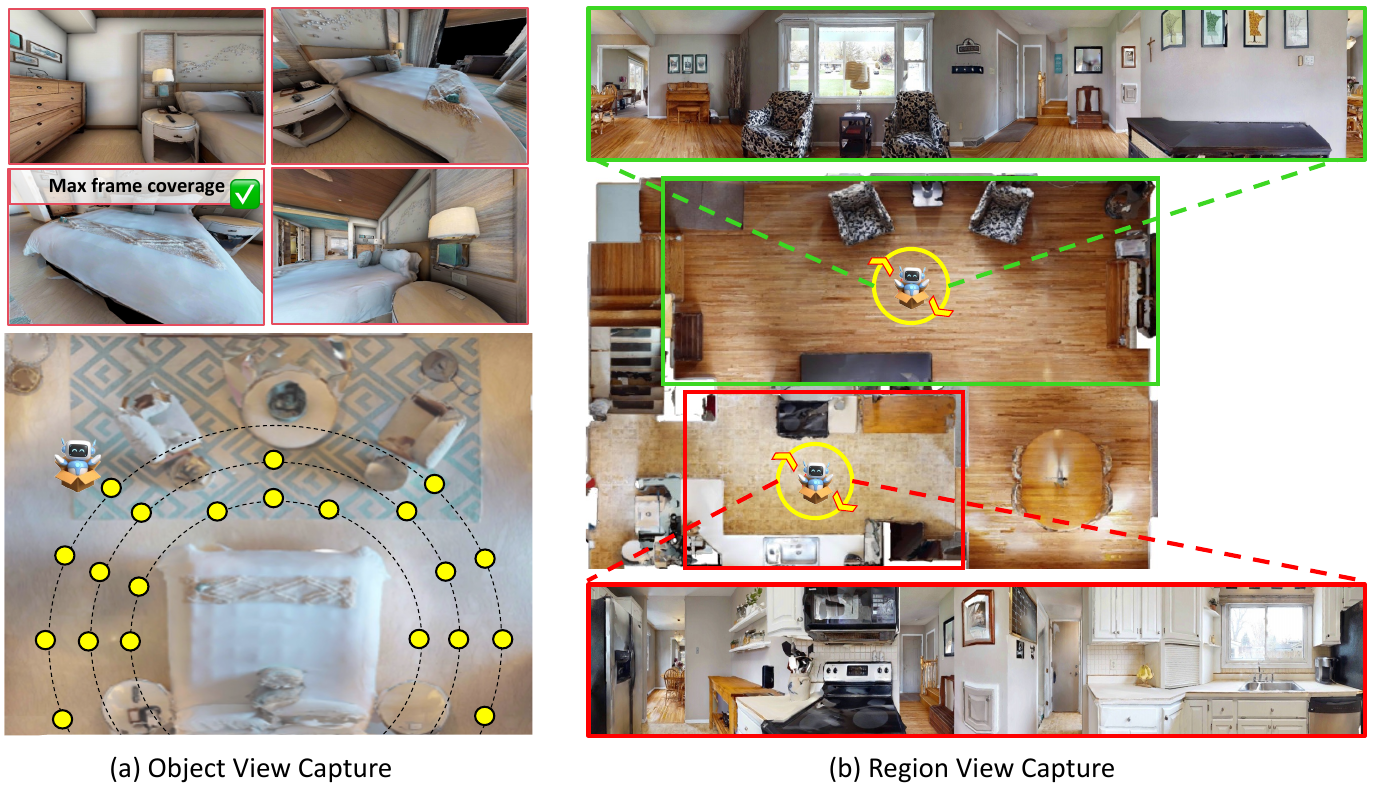}
\caption{
Representative object-view and region-view capture.
}
\label{fig:object_view_capture}
\end{figure*}

As shown in Figure~\ref{fig:object_view_capture}(a), for each object instance, we sample navigable candidate viewpoints at $10^\circ$ intervals within $r \in [0.5\,\mathrm{m}, 3\,\mathrm{m}]$. At each viewpoint, the camera is oriented toward the object centroid, and the view with the highest visible coverage is selected as the representative object view. This favors views in which the target instance is more complete.
As shown in Figure~\ref{fig:object_view_capture}(b), for each region, we approximate a pseudo-center as the midpoint of the bounding box enclosing all objects in the region. We then capture a panoramic observation at this pseudo-center, providing annotators with a broad view of the region layout, dominant room type, and surrounding context. The resulting region view supports room-category labeling and discriminative region-description annotation.

\subsection{Object Filtering}
After inspecting object labels and views, we remove categories that are overly abstract, non-countable, difficult to define as stable navigation targets, or frequently associated with low-quality views, such as beam, coat hanger, cable, sponge, toothpaste, socket, shoe, knob, and socks.
During annotation, object instances with low-quality views are also marked and excluded from instance-level navigation tasks.
After contrastive annotation and before task generation, we further exclude object instances without reliable forward-facing visibility or valid navigable viewpoints. This follows common embodied navigation protocols, where the action space does not include look-up or look-down actions.

\subsection{Annotation Protocol and Quality Control}
Annotators write natural, visually grounded descriptions that uniquely identify each target region or object instance within its scene.
For both region and instance annotation, annotators use a contrastive comparison interface that displays the target and its same-category distractors.
For region annotation, the interface shows the target region panorama and same-room-category distractor regions, each with views and labels of its contained objects.
For instance annotation, the interface first shows all object views from the same base category within the scene, together with their identifiers. Each instance is then displayed with its object category, associated region identifier and category, region panorama, and verified discriminative region descriptions.
Annotators generally compare all same-category instances within the scene.
For dense categories such as cabinets, verified region context narrows comparisons to within-region candidates, while region cues in the descriptions preserve scene-level disambiguation.
Annotators may also inspect the 3D scene files when needed.

Each target region and object instance receives concise and detailed descriptions.
Concise descriptions capture the most salient discriminative cue, generally within seven words including the category label. Detailed descriptions add visual attributes, nearby objects, and spatial relations.
During cross-checking, reviewers inspect descriptions alongside target views, metadata, and distractors in the annotation interface. They check correctness, visual grounding, label consistency, and discriminability, revising descriptions with incorrect attributes, insufficient discriminative cues, or ambiguity among candidates.

\begin{figure*}[t!]
\centering
\includegraphics[width=1\linewidth]{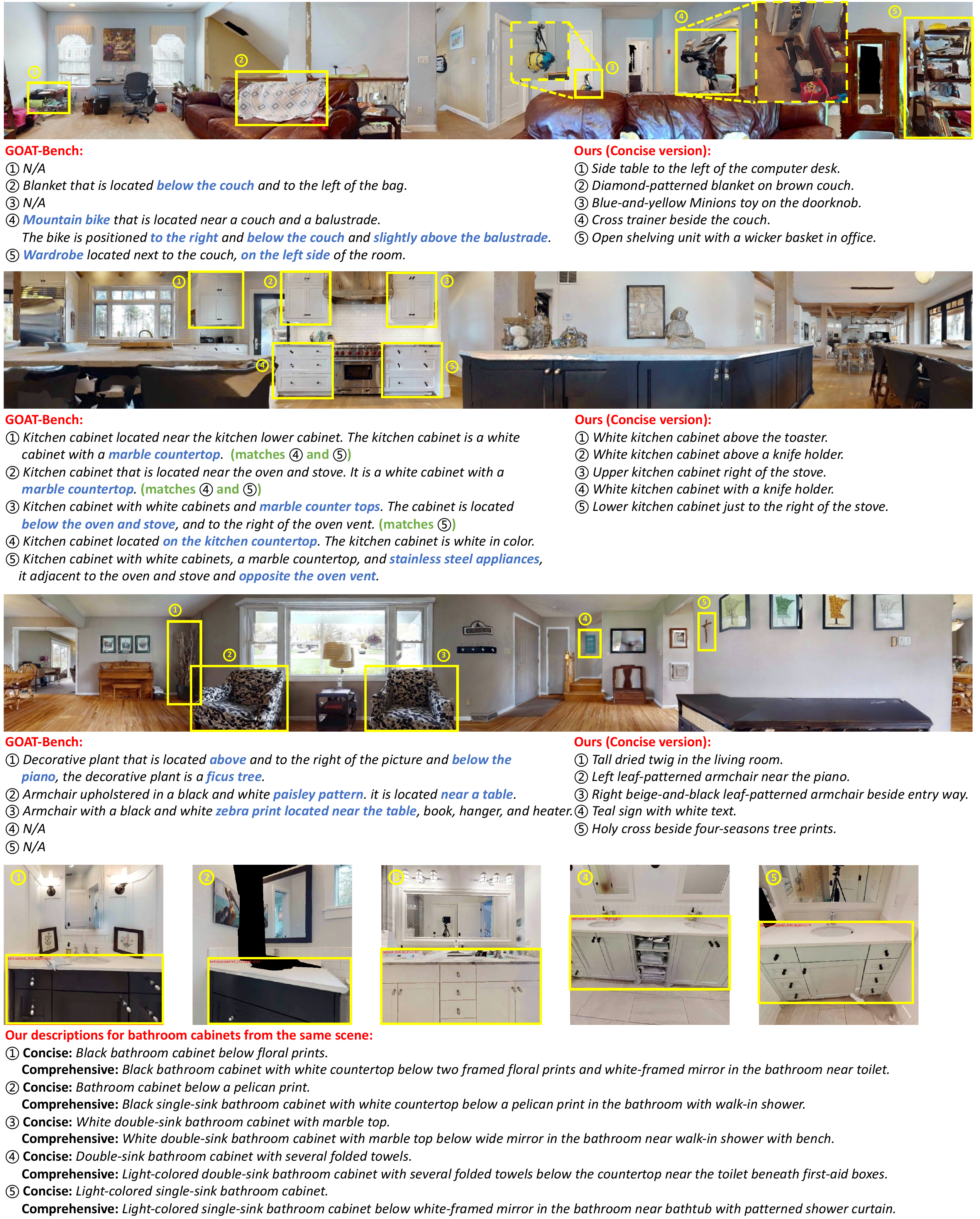}
\caption{
Comparison of instance-level descriptions in GOAT-Bench and \ourbenchmark.
Blue marks \textcolor{myblue}{\textbf{semantic errors}} and green marks \textcolor{mygreen}{\textbf{ambiguous descriptions}} that match multiple objects.
Our human-verified descriptions provide correct, fine-grained semantics and uniquely refer to the target instance in the scene.}\label{fig:goat_bench_error}
\end{figure*}

\begin{figure*}[t!]
\centering
\includegraphics[width=1.0\linewidth]{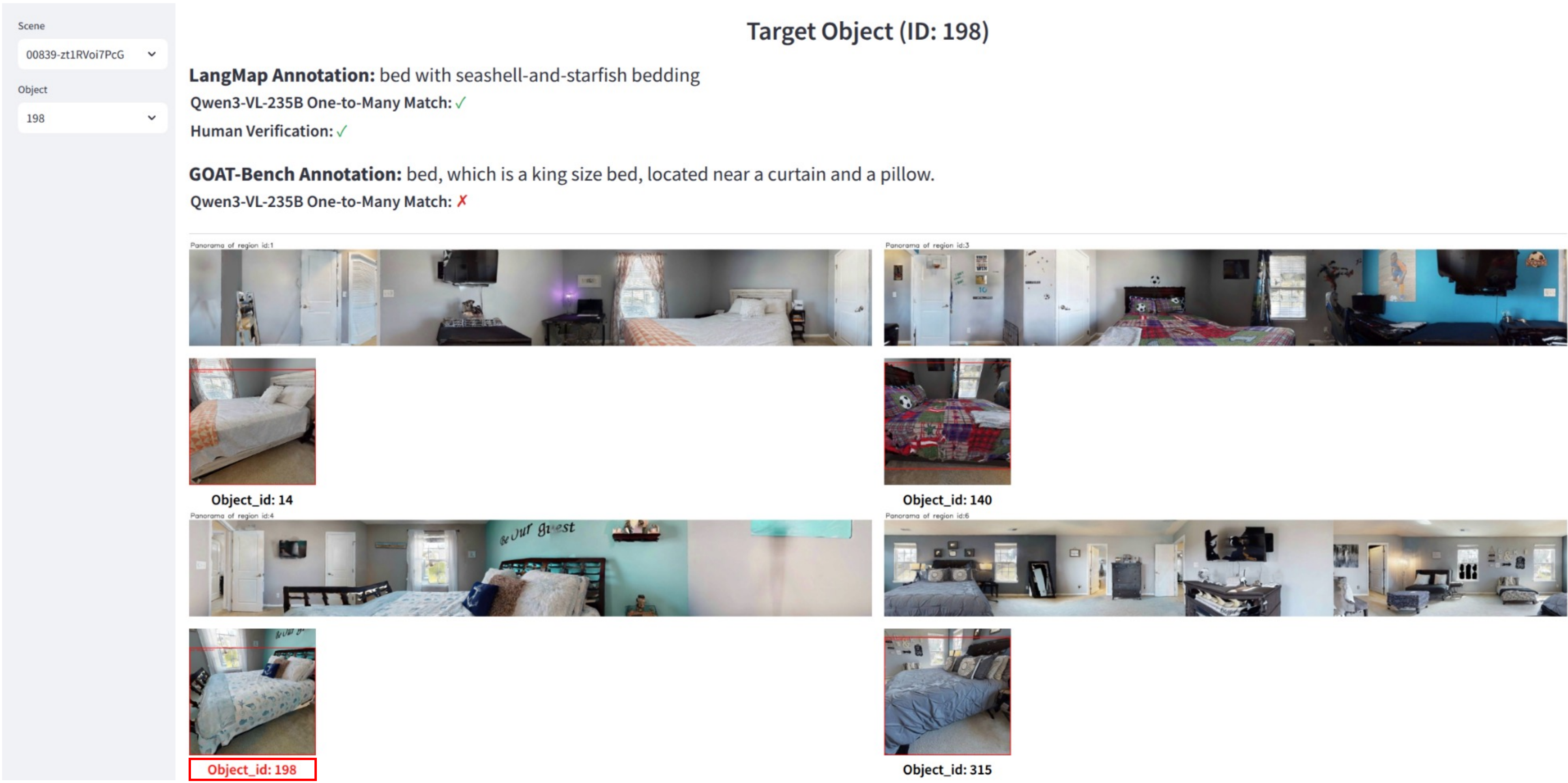}
\captionof{figure}{
Our annotation comparison viewer for qualitative analysis. It displays the target and same-category distractors to assess whether each description uniquely identifies the target.
}
\label{fig:contrast_view}
\end{figure*}

\section{Qualitative Comparison of Annotations in GOAT-Bench and \ourbenchmark}

We provide a Streamlit-based interactive viewer for directly comparing instance annotations from GOAT-Bench~\cite{goatbench} and \ourbenchmark.
As shown in Figure~\ref{fig:contrast_view}, the viewer allows users to select scenes and object instances, and displays the target object, same-category object crops, and corresponding region panoramas for side-by-side inspection.
Figure~\ref{fig:goat_bench_error} provides further qualitative comparisons of instance descriptions across regions, where we highlight \textcolor{myblue}{\textbf{semantic errors}} (\eg, category, attribute, or relation) in blue and \textcolor{mygreen}{\textbf{ambiguous descriptions}} that match multiple object instances in green.
These examples show that VLM-generated descriptions in GOAT-Bench often contain semantic inaccuracies or insufficiently discriminative cues.
In contrast, the human-verified descriptions in \ourbenchmark\ are concise, accurate, and more discriminative, supporting more reliable evaluation of language-conditioned navigation.

\section{Failure Visualization.}
\begin{figure*}[h!]
\centering
\includegraphics[width=1.0\linewidth]{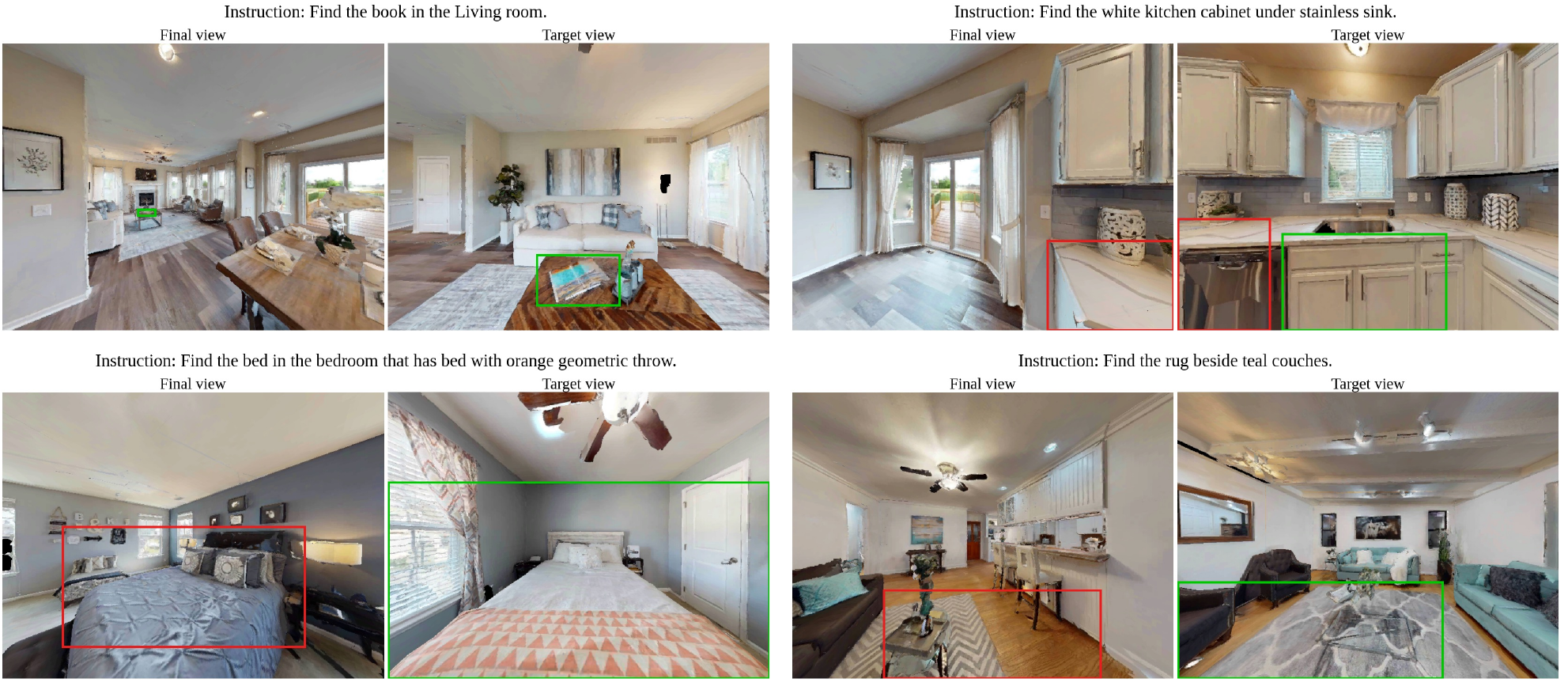}
\caption{Failure cases of \ourmodel. Green boxes mark ground-truth targets and red boxes mark distractors at the stop location.}
\label{fig:failure_vis}
\end{figure*}

Figure~\ref{fig:failure_vis} shows four representative failures of \ourmodel.
The top-left case illustrates premature stopping despite target visibility. The remaining cases show failures to distinguish targets using visual attributes (bottom row) and spatial relations (top right). These failures highlight the need for precise language grounding and reliable termination.

\section{Additional Details of Bounded Diverse Memory}
\label{sec:supp_bdm}

Algorithm~\ref{alg:memory} summarizes \ourmemory's update and retrieval procedures.
The online update maintains a bounded set of RGB memory states by pruning temporally redundant entries.
Retrieval anchors the latest state and iteratively selects memories to promote global semantic diversity.
Selected memories are passed to the high-level planner for waypoint and heading selection.
To reach the selected waypoint, the agent replays the recorded trajectory after pruning intermediate waypoints between spatially close positions, without querying the simulator pathfinder.

\begin{algorithm}[h]
\scriptsize
\caption{\ourmemory}
\label{alg:memory}
\begin{algorithmic}[1]
\State \textbf{Input:} Current state $\mathbf{x}_t=(s_t,\mathbf{I}_t,\mathbf{e}_t,\mathbf{p}_t)$, storage budget $N_{\max}$, retrieval budget $K$

\State \textbf{Global State:}
$\mathcal{M}=\{(\mathbf{x}_{s_i},\mathcal{P}_{i\to i+1})\}_{i=0}^{n-1}$,
$n\le N_{\max}$
\State \quad where
$\mathcal{P}_{i\to i+1}=\{\mathbf{p}_k\}_{k=s_i}^{s_{i+1}-1}$
and
$\mathcal{P}_{i\to j} \triangleq \text{Concat}\big(\{\mathcal{P}_{k\to k+1}\}_{k=i}^{j-1}\big)$

\vspace{0.6em}
\Function{Update}{$\mathbf{x}_t$}
  \State $\mathcal{M} \gets \mathcal{M} \cup \{(\mathbf{x}_{t}, \{\mathbf{p}_{t}\})\}$
  \State $n \gets n+1$
  \If{$n > N_{\max}$}
    \State $i^* \gets \arg\min_{0<i<n-1}(s_i-s_{i-1})$
    \State $\mathcal{P}_{i^*-1\to i^*+1} \gets \text{Concat}(\mathcal{P}_{i^*-1\to i^*}, \mathcal{P}_{i^*\to i^*+1})$
    \State $\mathcal{M} \gets \mathcal{M} \setminus \{(\mathbf{x}_{s_{i^*}}, \mathcal{P}_{i^*\to i^*+1})\}$
    \State $n \gets n-1$
\EndIf
\EndFunction

\vspace{0.6em}
\Function{Retrieve}{$\mathcal{M}$, $K$}
    \State $\mathcal{R} \leftarrow \{n-1\}$
    \While{$|\mathcal{R}| < \min(K, n)$}
        \State $i^* \leftarrow
        \arg\min\limits_{i \notin \mathcal{R}}
        \max\limits_{j \in \mathcal{R}}
        \langle \hat{\mathbf{e}}_{s_i}, \hat{\mathbf{e}}_{s_j} \rangle$
        \State $\mathcal{R} \leftarrow \mathcal{R} \cup \{i^*\}$
    \EndWhile
    \State \Return $\{(\mathbf{x}_{s_i}, \mathcal{P}_{i\to n-1})\}_{i \in \mathcal{R}}$ sorted by $s_i$
\EndFunction

\end{algorithmic}
\end{algorithm}

\section{Additional Results}

\subsection{SeqSR Results.}
Table~\ref{tab:seqsr} reports SeqSR@1--5 results on \ourbenchmark.
Across all methods, SeqSR decreases sharply as $k$ increases, showing that completing multi-goal sequences remains challenging.

\begin{table}[h!]
\centering
\small
\setlength{\tabcolsep}{3pt}
\caption{
SeqSR@1--5 Results on \ourbenchmark.
}
\begin{tabular}{l|c|c|c|c|c}
\toprule[1.5pt]
Method & SeqSR@1 & SeqSR@2 & SeqSR@3 & SeqSR@4 & SeqSR@5 \\
\midrule
PSL & 11.9 & 3.1 & 0.7 & 0.3 & 0.0 \\
SenseAct-M & 6.0 & 1.0 & 0.1 & 0.1 & 0.1 \\
3D-Mem-3B & 6.9 & 1.1 & 0.3 & 0.0 & 0.0 \\
3D-Mem-7B & 12.4 & 5.1 & 1.9 & 0.8 & 0.1 \\
MTU3D & 25.0 & 11.0 & 5.4 & 2.5 & 1.1 \\
Uni-NaVid & \cellcolor{teal!15}\textbf{27.1} & 10.4 & 3.5 & 1.4 & 0.6 \\
\midrule
\textbf{\ourmodel-3B} & 26.4 & 14.4 & 7.1 & \cellcolor{teal!15}\textbf{3.8} & \cellcolor{teal!15}\textbf{2.9} \\
\textbf{\ourmodel-7B} & 26.8 & \cellcolor{teal!15}\textbf{14.6} & \cellcolor{teal!15}\textbf{7.8} & \cellcolor{teal!15}\textbf{3.8} & 1.7 \\
\bottomrule[1.5pt]
\end{tabular}
\label{tab:seqsr}
\end{table}

\subsection{Ablation of $N_{max}$ and $K$.}
We study the impact of the memory storage budget $N_{\max}$ and retrieval budget $K$. 
As shown in Table~\ref{tab:ablation_mem_size}, both factors affect performance.
Latest Memory underperforms with limited long-term context, and Full Memory improves performance at the cost of substantial redundancy.
For our memory, a smaller $N_{\max}$ reduces coverage of past observations, a smaller $K$ limits the semantic diversity of retrieved context, and performance saturates beyond $N_{\max}{=}50$.
Despite using a much smaller budget, \ourmemory\ outperforms Full Memory, supporting the efficacy of global-uniform update and semantically diverse retrieval.

\begin{table}[h!]
\centering
\small
\setlength{\tabcolsep}{4pt}
\caption{
Ablation of $N_{\max}$ and $K$. Ours uses Bounded Diverse Memory.
}
\begin{tabular}{lcc|ccc}
\toprule[1.5pt]
&$N_{\max}$ & $K$ & SR & SeqSR@2 & SPL \\
\midrule
Latest Memory & 50 & 10 & 36.4 & 11.4 & 14.9 \\ 
Full Memory & 656 & 10 &40.4& 13.2& 16.9 \\ 
\midrule
\multirow{5}{*}{\textbf{Ours}}&25 & 10 &  41.5 &  13.8  & 17.3 \\ 
&50 & 5 &  40.0 & 12.2 & 17.0 \\
&50 & 10 & 42.8 & 14.6 & 18.1 \\
& 75 & 10 & 42.3 & 13.8 & 17.7 \\
& 100 & 10 & 42.6 & 13.6 & 17.6 \\ 
\bottomrule[1.5pt]
\end{tabular}
\label{tab:ablation_mem_size}
\end{table}

\subsection{Bootstrap Confidence Intervals}
We report 95\% confidence intervals from 10,000 scene-cluster bootstrap resamples, using the same resamples across methods and retaining complete episodes for SeqSR.
As shown in Table~\ref{tab:multigoal_ci}, \ourmodel\ achieves top-tier performance.
Compared with Uni-NaVid, \ourmodel-7B improves multi-goal SR by 8.4 points [6.5, 10.4] and SPL by 3.1 points [1.7, 4.4] (95\% paired CIs).
Against MTU3D$^{*}$ with matched HFOV, it improves SR by 4.8 points [2.2, 7.3].

\begin{table}[h]
\centering
\small
\setlength{\tabcolsep}{4pt}
\caption{Multi-goal results (\%) with 95\% scene-cluster bootstrap CIs.
$^{*}$: MTU3D with a horizontal field of view matched to \ourmodel.}
\label{tab:multigoal_ci}
\begin{tabular}{c|l|ccc}
\toprule[1.5pt]
Config & Method
& SR$\uparrow$ & SeqSR@2$\uparrow$ & SPL$\uparrow$ \\
\midrule
\multirow{4}{*}{RGB-D,Mask}
& 3D-Mem-3B
& 13.5 [11.8, 15.2] & 1.1 [0.4, 1.9] & 6.4 [4.9, 8.1] \\
& 3D-Mem-7B
& 30.1 [26.6, 33.7] & 5.1 [3.1, 7.4] & 17.3 [15.0, 19.8] \\
& MTU3D
& 41.2 [37.8, 44.7] & 11.0 [8.1, 14.0] & 24.3 [21.9, 26.7] \\
& MTU3D$^{*}$
& 38.0 [34.5, 41.4] & 7.9 [5.4, 10.6] & 21.0 [18.8, 23.1] \\
\midrule
\multirow{3}{*}{RGB}
& Uni-NaVid
& 34.4 [31.0, 37.6] & 10.4 [7.9, 12.9] & 15.0 [13.3, 16.7] \\
& \textbf{\ourmodel-3B}
& 41.5 [37.9, 44.9] & 14.4 [11.1, 17.9] & 17.5 [15.5, 19.5] \\
& \textbf{\ourmodel-7B}
& 42.8 [39.3, 46.1] & 14.6 [11.2, 18.2] & 18.1 [16.1, 20.0] \\
\bottomrule[1.5pt]
\end{tabular}%
\end{table}

\subsection{Visualization of Multi-Goal Navigation.}
\begin{figure}[h!]
\centering
\includegraphics[width=1.\textwidth]{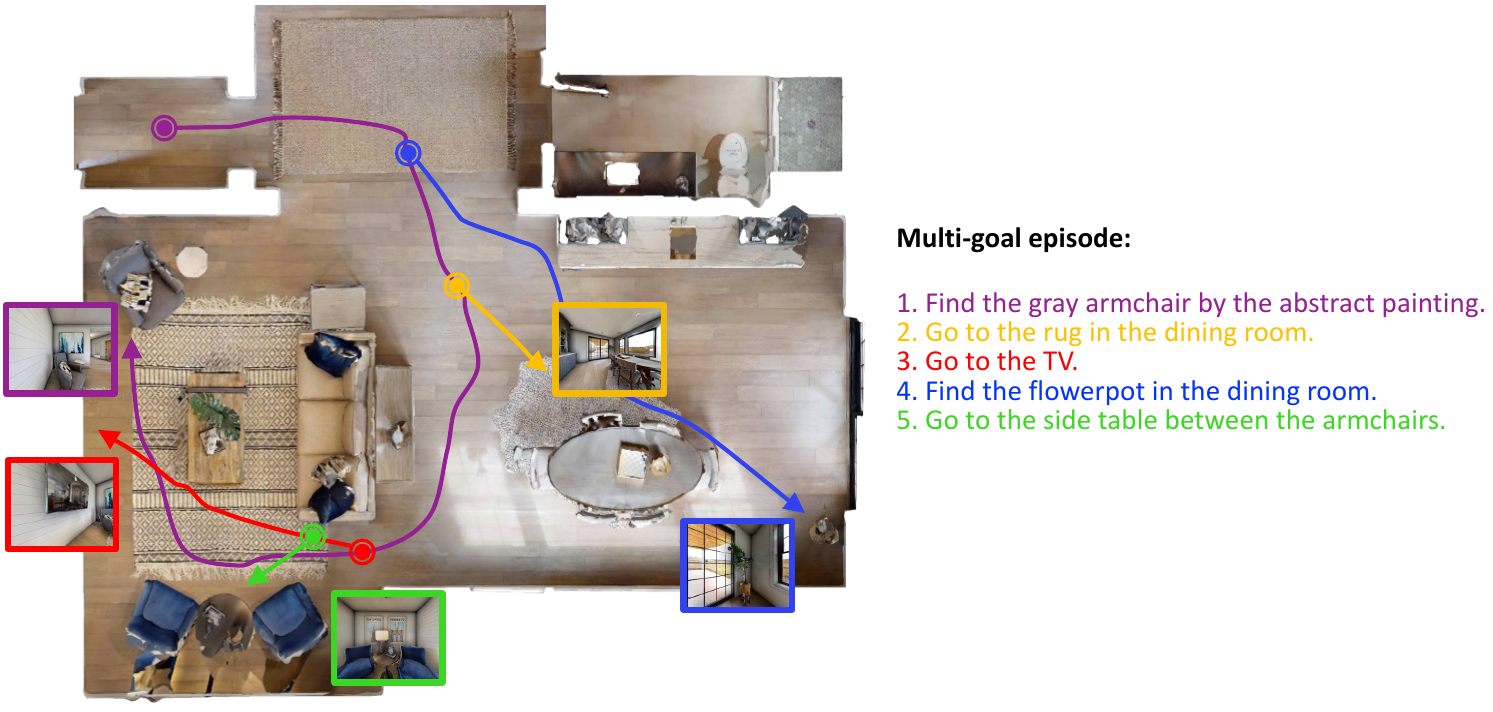}
\caption{
Example of sequential multi-goal navigation.
The colored paths denote different goals.
}\label{fig:multigoal_example}
\end{figure}

Figure~\ref{fig:multigoal_example} shows a multi-goal episode and its target views.

\end{document}